\pdfoutput=1

\documentclass[11pt]{article}

\usepackage{naacl2021}
\usepackage{times}
\usepackage{latexsym}
\usepackage{color}

\newcommand{\ind}{\perp\!\!\!\!\perp}
\newcommand{\RNum}[1]{\uppercase\expandafter{\romannumeral #1\relax}}

\usepackage[T1]{fontenc}

\usepackage[utf8]{inputenc} 

\usepackage{microtype}
\usepackage{graphicx}
\usepackage{amsmath}
\usepackage{multirow}
\usepackage{booktabs} 
\usepackage{amsfonts}
\usepackage{placeins}

\usepackage{algorithm}
\usepackage{algorithmic}

\definecolor{c1}{RGB}{169,199,236}
\definecolor{c2}{RGB}{146,201,235}
\definecolor{c3}{RGB}{65,137,221}
\definecolor{c4}{RGB}{0,38,127}
\definecolor{c5}{RGB}{0,32,78}

\usepackage[symbol]{footmisc}

\title{Everything Has a Cause: Leveraging Causal Inference in \\ Legal Text Analysis}

\author{Xiao Liu$^{1}$ \\
  Affiliation / Address line 1 \\
  Affiliation / Address line 2 \\
  Affiliation / Address line 3 \\
  \texttt{email@domain} \\\And
  Second Author \\
  Affiliation / Address line 1 \\
  Affiliation / Address line 2 \\
  Affiliation / Address line 3 \\
  \texttt{email@domain} \\}
\author{
	Xiao Liu$^{1}$\footnotemark[1], 
	Da Yin$^{2}$\footnotemark[1],
	Yansong Feng$^{1,3}$\footnotemark[2],  
	Yuting Wu$^{1}$ \and
	Dongyan Zhao$^{1,3}$ \\
	$^1$Wangxuan Institute of Computer Technology, Peking University, China\\
	$^2$Computer Science Department, University of California, Los Angeles\\
	$^3$The MOE Key Laboratory of Computational Linguistics, Peking University, China\\
	{\tt \{lxlisa,fengyansong,wyting,zhaody\}@pku.edu.cn} \\
	{\tt da.yin@cs.ucla.edu}\\
}

\date{}

\begin{document}
\maketitle
\footnotetext[1]{\;\;Equal contribution.}
\footnotetext[2]{\;\;Corresponding author.}

\begin{abstract}
Causal inference is the process of capturing cause-effect relationship among variables. 
Most existing works focus on dealing with structured data, while mining causal relationship among factors from unstructured data, like text, 
has been less examined, but is of great importance, especially in the legal domain. 
In this paper, we propose a novel \textbf{G}raph-based \textbf{C}ausal \textbf{I}nference (\textit{GCI}) framework, which builds causal graphs from fact descriptions without much human involvement and enables causal inference to facilitate legal practitioners to make proper decisions. We evaluate the framework on a challenging \emph{similar charge disambiguation} task. 
Experimental results show that \textit{GCI} can capture the nuance from fact descriptions among multiple confusing charges and provide explainable discrimination, especially in few-shot settings. We also observe that the causal knowledge contained in \textit{GCI} can be effectively injected into powerful neural networks for better performance and interpretability.   
Code and data are available at \url{ https://github.com/xxxiaol/GCI/}.
\end{abstract}
\section{Introduction}
\label{intro}
Causal inference is the process of exploring how changes on variable $T$ affect another variable $Y$. Here we call $T$ and $Y$ as \textbf{\emph{treatment}} and \textbf{\emph{outcome}}, respectively, and the changes on $T$ are called \textbf{\emph{intervention}}. In other words, the process of drawing a conclusion about whether and how $Y$ changes when intervening on $T$ is called \textbf{\emph{causal inference}}.

Most research in causal inference is devoted to analyzing structured data. Take the research question \textit{how smoking causes lung cancer}~\citep{pearl2018book} as an example. \textit{Smoking}, \textit{lung cancer}, together with distractors like \textit{age} are extracted from structured data, like electronic health records, and considered as factors. Usually, such studies properly organize those factors into human-designed structures, e.g., a causal directed acyclic graph~\citep{Wright1921Correlation} with factors \{\textit{smoking}, \textit{age}, \textit{lung cancer}\} as  nodes and causal relations \{\textit{smoking $\rightarrow$ lung cancer}, \textit{age $\rightarrow$ lung cancer}\} as edges, and perform inference on such structures.

Recent works attempt to integrate text information into causal inference~\citep{egami2018make,veitch2019using,yao2019estimation,keith2020text}, but they mainly treat the text as a single node in the causal graph, which is relatively coarse-grained. For instance, \citet{yao2019estimation} investigate how the complaints from consumers affect the company's responses (admission or denial). They regard the entire text of complaint as a treatment, without looking into 
different aspects of the text, like the events that consumers complained about and the compensation that consumers requested.

\begin{figure}
    \centering
    \includegraphics[width=0.95\linewidth]{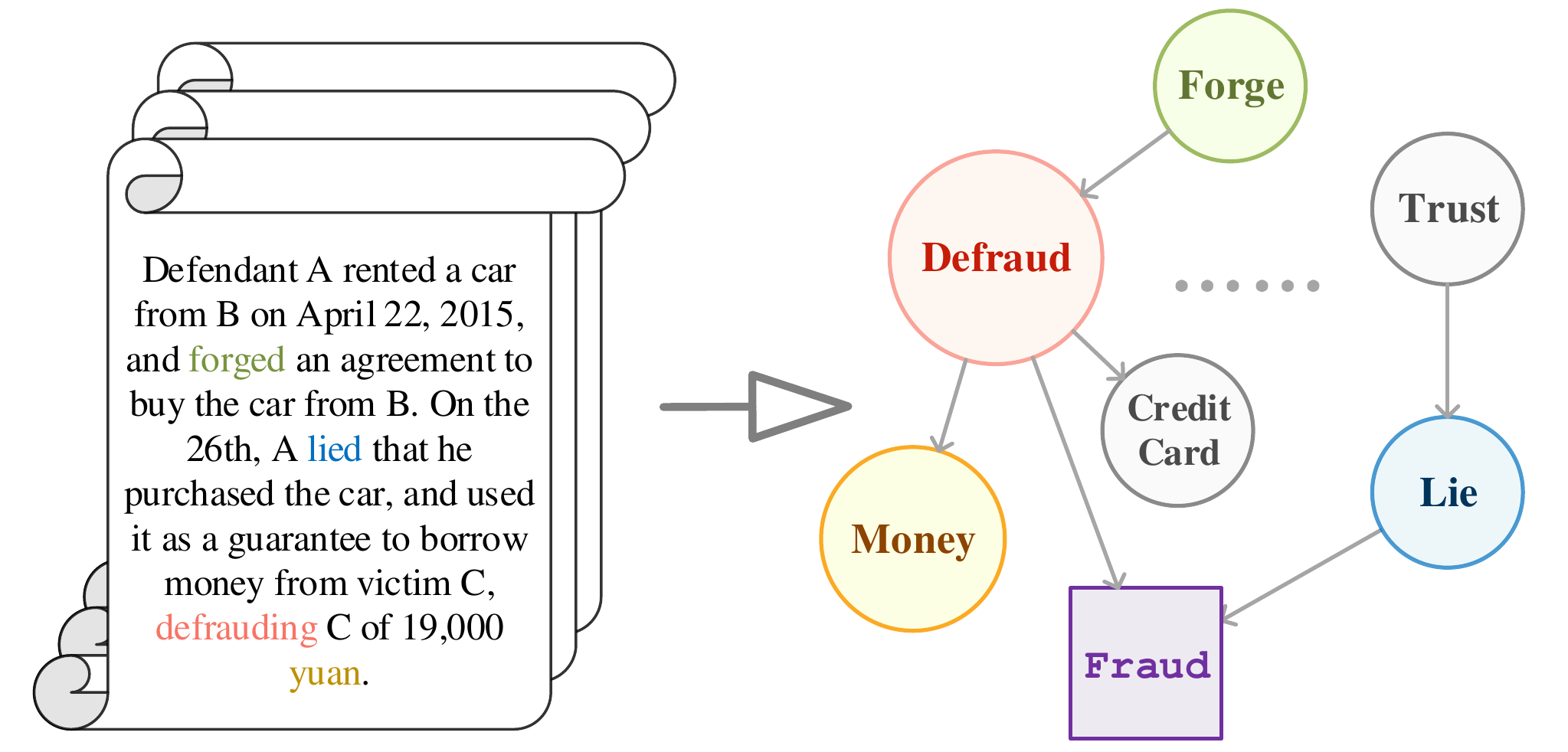}
    \caption{An example of generated causal graph for the charge \texttt{fraud}. Colored words are matched between the exemplified fact description and the graph.}
    \label{fig-intro}
\end{figure}

Actually, discovering causal relationship \emph{inside} unstructured and high-dimensional data, like text, is also beneficial or even crucial for scenarios involving reading comprehensive text and making decisions accordingly. For instance, 
when a legal AI system assists judges to deal with complicated cases that involve multiple parties and complex events, causal inference could help to figure out the exact distinguishable elements that are crucial for fair and impartial judgements.
As shown in Figure~\ref{fig-intro}, if the system can automatically spot two essential key points, 1) the deceitful acts of the defendant, and 2) obtaining properties from the victim, from the unstructured fact descriptions, then the prediction \texttt{fraud} can be more convincing and helpful, rather than a label from a black box.
In practice, 
we would expect a legal AI system to provide human-readable and sound explanations to help the court make the right decisions. 
It is worthwhile especially for underdeveloped areas, where such techniques could help the judges of rural areas with more trustworthy references from previous judgements. This would further 
help to maintain the principle of \textit{treating like cases alike} in many continental law systems. A main challenge in judicial practice is to distinguish between similar charges, we thus propose a new task \textbf{\emph{similar charge disambiguation}} as a testbed. Cases of similar charges often share similar context, and a system is expected to mine the nuance of reasoning process from the text. 

However, performing causal inference on fact descriptions of criminal cases is not trivial at all. It poses the following challenges.
1) Without expert involvement, it is not easy to extract factors that are key to prediction, and organize them in a reasonable form that can both facilitate the inference process and tolerate noise. For example, automatically extracted elements may not cover all the key points in the textual descriptions, and automatically built graphs may contain unreliable edges.
2) It is not easy to benefit from both traditional causal inference models and modern neural architectures.

In this paper, we propose a novel \textbf{G}raph-based \textbf{C}ausal \textbf{I}nference (\textit{GCI}) framework, which could effectively apply causal inference to legal text analysis. \textit{GCI} first recognizes key factors by extracting keywords from the fact descriptions and clustering similar ones into groups as individual nodes. Then we build causal graphs on these nodes with a causal discovery algorithm that can tolerate unobserved variables, reducing the impact of missing elements.
We further estimate the causal strength of each edge, to weaken unreliable edges as much as possible, and apply the refined graph to help decision making.
Experimental results show that our \textit{GCI} framework can induce reasonable causal graphs and capture the nuance from plain text for legal applications, especially with few training data. 

We also explore the potential of \textit{GCI} by integrating the captured causal knowledge into neural network (NN) models.
We propose two approaches: 1) imposing the causal strength constraints to the NN's attention weights; 2) applying recurrent neural networks on the causal chains extracted from our causal graphs. 
Experiments indicate that our methods can successfully inject the extracted causal knowledge and thus enhance the NN models.
We also show that integrating \textit{GCI} helps to mitigate the underlying bias in data towards the final prediction.

Our main contributions are as follows:
1) We propose a novel graph-based causal inference (\textit{GCI}) framework to 
apply causal inference to unstructured text, without much human involvement.
2) We explore to equip popular neural network models with our \textit{GCI}, by encouraging neural models to learn from causal knowledge derived from \textit{GCI}.
3) We evaluate our methods on a legal text analysis task, similar charge disambiguation, and experimental results show that our \textit{GCI} can capture the nuance from plain fact descriptions, and further help improve neural models with interpretability.  


\section{Background}
\label{sec:background}
Two types of questions are typically related to causality. The first is whether there is causal relationship between a set of variables, and the second question is when two variables $T$ and $Y$ are causally related, how much would $Y$ change if we change the value of $T$. Both of them are discussed in our \emph{GCI} framework, and we first briefly introduce the key concepts.

\subsection{Causal Discovery} 
Causal discovery corresponds to the first type of questions.
From the view of graph, causal discovery requires models to infer causal graphs from observational data. 
In our \emph{GCI} framework, we leverage \textbf{\emph{Greedy Fast Causal Inference}} (GFCI) algorithm~\citep{ogarrio2016hybrid} to implement causal discovery. 
GFCI combines score-based and constraint-based algorithms. It merges the best of both worlds, performing well like score-based methods, and not making too unrealistic assumptions like constraint-based ones. Specifically, GFCI does not rely on the assumption of no latent confounders, thus is suitable in our situation. More details of GFCI algorithm and its advantage is provided in Appendix~\ref{algo}.

The output of GFCI is a graphical object called \textbf{\emph{Partial Ancestral Graph}} (PAG). PAG is a mixed graph containing the features common to all Directed Acyclic Graphs (DAGs) that represent the same conditional independence relationship over the measured variables. In other words, PAG entails all the possibilities of valid DAGs concerning the original data. In causal inference settings~\cite{zhang2008causal}, four types of edges are provided in PAG, as listed in Table~\ref{table-pag}. With PAG, we are able to consider unobserved confounders and the uncertainty in causal inference.

\begin{table}[t!]
	\centering
    \scalebox{0.95}{
	\scriptsize
	\begin{tabular}{c|c}
		\toprule
		{\textbf{Edge}} & \textbf{Meaning} \\
		\midrule
		A $\:\rightarrow\:$ B & A causes B. \\
		A $\:\leftrightarrow\:$ B & There is an unobserved confounder of A and B.\\
		A $\circ\!\!\rightarrow$ B & Either A causes B, or unobserved confounder.\\
		A $\circ\!\!-\!\!\circ$ B & Either A causes B, or B causes A, or unobserved confounder.\\
		\bottomrule
	\end{tabular}
	}
	\caption{Summary of edge types in PAG.}
	\label{table-pag}
\end{table}

\subsection{Causal Strength Estimation}
\label{cse}
Causal strength estimation deals with the second type of questions. It is the very task to quantify the causal strength of each learned relation, i.e., whether the relation is strong or weak. To precisely estimate causal strength, confounders need to be kept the same. \textbf{\emph{Confounder}} is a variable causally influencing both treatment $T$ and outcome $Y$.
Take the example of \emph{smoking, lung cancer} and \emph{age} in Section \ref{intro}. 
Here we study if there is causal relationship between \emph{smoking} ($T$) and \emph{lung cancer} ($Y$), and \emph{age} is a confounder $C$. It is straightforward to compare the proportion of \emph{lung cancer} among smokers and non-smokers.
However, \emph{age} influences both \emph{smoking} and \emph{lung cancer}. Older people are more likely to smoke. They also have a much higher risk of suffering from cancer. If we do not consider the value of \emph{age}, its influence to \emph{lung cancer} will be regarded as \emph{smoking}'s influence, thus wrongly amplify the causal effect of \emph{smoking} to \emph{lung cancer}.

In our \emph{GCI} framework, we apply \textbf{\emph{Average Treatment Effect}} (ATE)~\cite{Holland1986StatisticsAC} as a measure of causal strength. All variables are binary in our work. So given an edge $T \rightarrow Y$, we quantify how the outcome $Y$ is expected to change if we modify the treatment $T$ from 0 to 1:
\begin{equation}
    \psi_{T,Y}=E[Y\mid do(T=1)]-E[Y\mid do(T=0)],
\end{equation}
where $E$ means expectation, and the do-calculus $do(T=1)$ indicates intervention on $T$, setting its value to $1$. 

We utilize the \textbf{\emph{Propensity Score Matching}} (PSM) method to estimate ATE. 
PSM finds the pairs of comparable samples with the most similar propensity scores, where each pair consists of one sample in the treated group and one in the untreated.
Given the great similarity between the two samples, we could make a direct comparison between them. 
Specifically, propensity score $L(z)=P(T=1\mid Z=z)$ is the probability of treatment being assigned to $1$ given a set of observed confounders $z$.
As $T$ is binary, we have $T \ind Z \mid L$ ($\ind$ means independence). So matching on propensity scores equals matching on the full set of confounders.

\section{Graph-based Causal Inference Framework}
\label{pure}
\begin{figure*}[t!]
    \centering
    \includegraphics[width=0.85\linewidth]{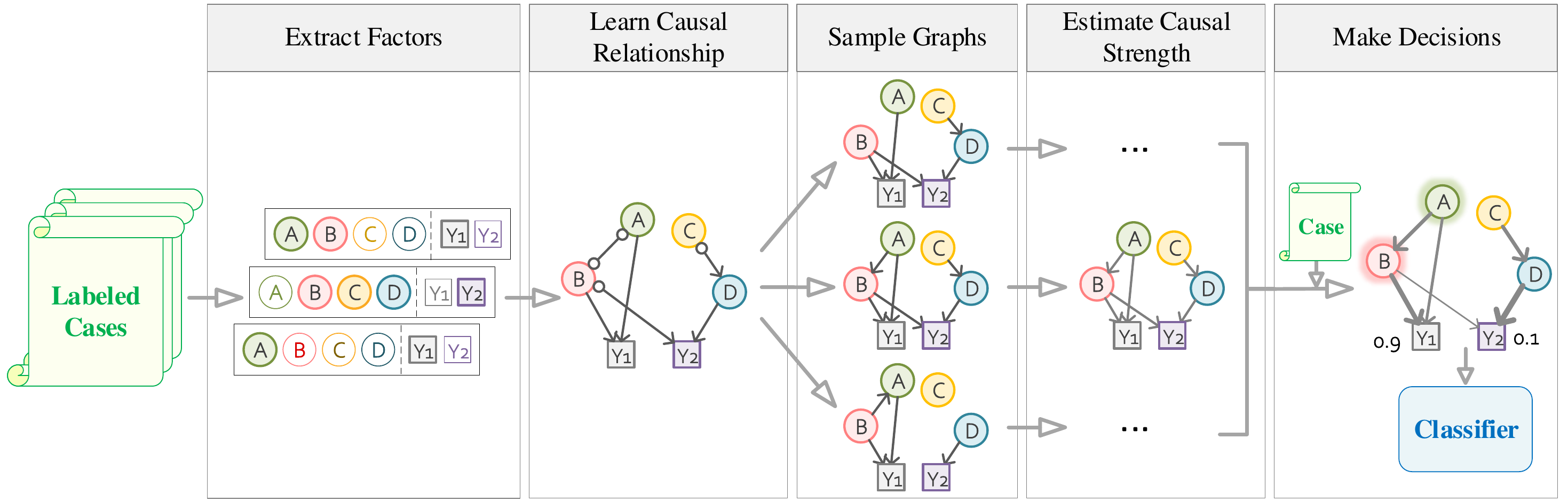}
    \caption{Overall architecture of \textit{GCI}. In the \textit{Extract Factors} phase, solid circles indicate that these factors exist in this case, while hollow circles mean the opposite. For the nodes of causal graphs, $A$, $B$, $C$, and $D$ denote the key points contributing to prediction, while $Y_1$ and $Y_2$ indicate the charges to discriminate. Four types of edges $\rightarrow$, $\leftrightarrow$, $\circ\!\!\rightarrow$, and $\circ\!\!-\!\!\circ$ exist in the PAG in the \textit{Learn Causal Relationship} phase, and they are converted to $\rightarrow$ edges in the sampled graphs. In the rightmost phase, shaded factors are matched between fact descriptions of cases and the graph.}
    \label{fig-pure}
\end{figure*}
Our graph-based causal inference (\textit{GCI}) framework consists of three parts, constructing the causal graph, estimating causal strength on it, and making decisions. 
Figure~\ref{fig-pure} shows the overall architecture.

\subsection{Task Definition}
We first define the \textit{similar charge disambiguation} task. Given the fact descriptions of criminal cases $\mathbb{D}=\{d_1,d_2,\ldots,d_N\}$, a system is expected to classify each case into one charge from the similar charge set $\mathbb{C}=\{c_1,c_2,\ldots,c_M\}$.

\subsection{Causal Graph Construction} 

\paragraph{Extracting Factors.}
To prepare nodes for the causal graph, we calculate the importance of word $w_j$ for charge $c_i$ using YAKE~\citep{campos2020yake}.
We enhance YAKE with inverse document frequency~\citep{jones1972statistical} to extract more discriminative words of each charge.

To discriminate the similar charges, we select $p$ words with the highest importance scores for each charge, cluster them into $q$ classes to merge similar keywords. The $q$ classes together with the $M$ charges form the nodes of the causal graph. All these factors are binary. When the graph is applied to a case, each factor is of value $1$ if it exists in this case, and $0$ if not. Unlike factors extracted by experts, automatically extracted keywords may be incomplete, resulting in unobserved confounders in causal discovery.

\paragraph{Learning Causal Relationship.}
The next step is to build edges for the graph, in other words, discover the causal relationship between different factors. 
To learn causal relations and tackle the unobserved confounder problem, we use GFCI~\citep{ogarrio2016hybrid}, which does not rely on the assumption of no unobserved confounders. As mentioned in Section \ref{sec:background}, the output of GFCI is a PAG. Appendix \ref{appendix-exp} gives an example of a generated PAG in detail.

We further introduce constraints to filter noisy edges.
First, as the judgement is made based on the fact description, we do not allow edges from charge nodes to other ones, e.g., an edge from \texttt{fraud} to \textit{lie} is prohibited.
Second, given that causes usually appear before effects in time~\citep{black1956cannot}, and fact descriptions in legal text are often written in the temporal order of events, we thus consider the chronological order of descriptions as temporal constraints to filter noisy edges. If factor $A$ appears after $B$ in most cases, we will not allow the edge from $A$ to $B$. Note that this constraint does not imply there is an edge from $B$ to $A$, as chronological order is not a sufficient condition of causality.

\paragraph{Sampling Causal Graphs.}
PAG contains uncertain relations shown in Table~\ref{table-pag}, which leaves challenges for quantification and further application. So we sample $Q$ causal graphs from PAG. Among the four edge types, $\rightarrow$ and $\leftrightarrow$ are clear: in each sampled graph, $\rightarrow$ edges are retained and $\leftrightarrow$ edges are removed (because they do not indicate causal relations between the two nodes). For $\circ\!\!\rightarrow$ edges, they have two possible choices: being kept (cause) and being removed (unobserved confounder). 
In the absence of true possibility, we simply keep an edge with $1/2$ probability, and remove it with another $1/2$. And for $\circ\!\!-\!\!\circ$ edges, we give $1/3$ probability for $\rightarrow$, $\leftarrow$, and no edge, respectively.
The quality of each sampled graph $G_q$ is measured by its fitness with data $\mathbf{X}$, where we use the Bayesian information criterion $\mathrm{BIC}(G_q, \mathbf{X})$ to estimate~\citep{schwarz1978estimating}.

\subsection{Strength Estimation on Causal Graphs}
\label{inference}
As the resulting graphs are noisy in nature, we estimate the strength of the learned causal relations to refine a sampled causal graph. We assign high strength to edges with strong causal effect, and near-zero strength to edges that do not indicate causal relations or with weak effect.
We regard the Average Treatment Effect $\psi_{T,Y}^G$ (ATE, Section~\ref{cse}) as the strength of $T \rightarrow Y$ in graph $G$, and utilize the Propensity Score Matching (PSM, Section~\ref{cse}) to measure it:
\begin{equation}
    \hat{\psi}_{T,Y}^G=[\sum_{i: t_i=1}(y_i-y_j)+\sum_{i: t_i=0}(y_j-y_i)]/N,
\end{equation}
where $j=\mathop{\mathrm{argmin}}\limits_{k:t_k \neq t_i}|L(z_i)-L(z_k)|$ means the most similar instance in the opposite group of $i$, and $t_i, y_i, z_i$ are the value of treatment, outcome and confounders of instance $i$, respectively. 

\subsection{Making Decisions}
When applying the sampled causal graphs to the similar charge disambiguation task, we simply extract factors and map the case description with the graph accordingly, and decide which charge in $\mathbb{C}$ is more appropriate to this case.
Firstly, we compute the overall causal strength of each factor $T_j$ to $Y_i$ among the $Q$ sampled causal graphs, where $Y_i$ represents whether charge $c_i$ is committed:
\begin{equation}
    \tilde{\psi}_{T_j,Y_i}=\sum_{q=1}^Q \mathrm{BIC}(G_q, \mathbf{X})\times \hat\psi^{G_q}_{T_j,Y_i},
\end{equation}
where $\hat\psi^{G_q}_{T_j,Y_i}$ is the measured causal strength in $G_q$, and is $0$ if edge $T_j \rightarrow Y_i$ does not exist in $G_q$.

For each case, we then map the text with the graphs, and calculate scores for each charge:
\begin{equation}
    S(Y_i)=\sum_{T_j \in Tr(Y_i)}\tilde{\psi}_{T_j,Y_i}\times \tau(T_j), i \in \{1,\ldots,M\},
\end{equation}
where $\tau(T_j)$ is a dummy variable indicating the presence of $T_j$ in this case, and $Tr(Y_i)$ is the set of treatments of $Y_i$ (from the view of graph, the nodes pointing to $Y_i$). 
The calculated scores are fed into a random forest classifier~\citep{ho1995random} to learn thresholds between the charges. More advanced classifiers can also be used.

\section{Integration of Causal Analysis and Neural Networks}
\label{NN}
\begin{figure}[t!]
    \centering
    \includegraphics[width=\linewidth]{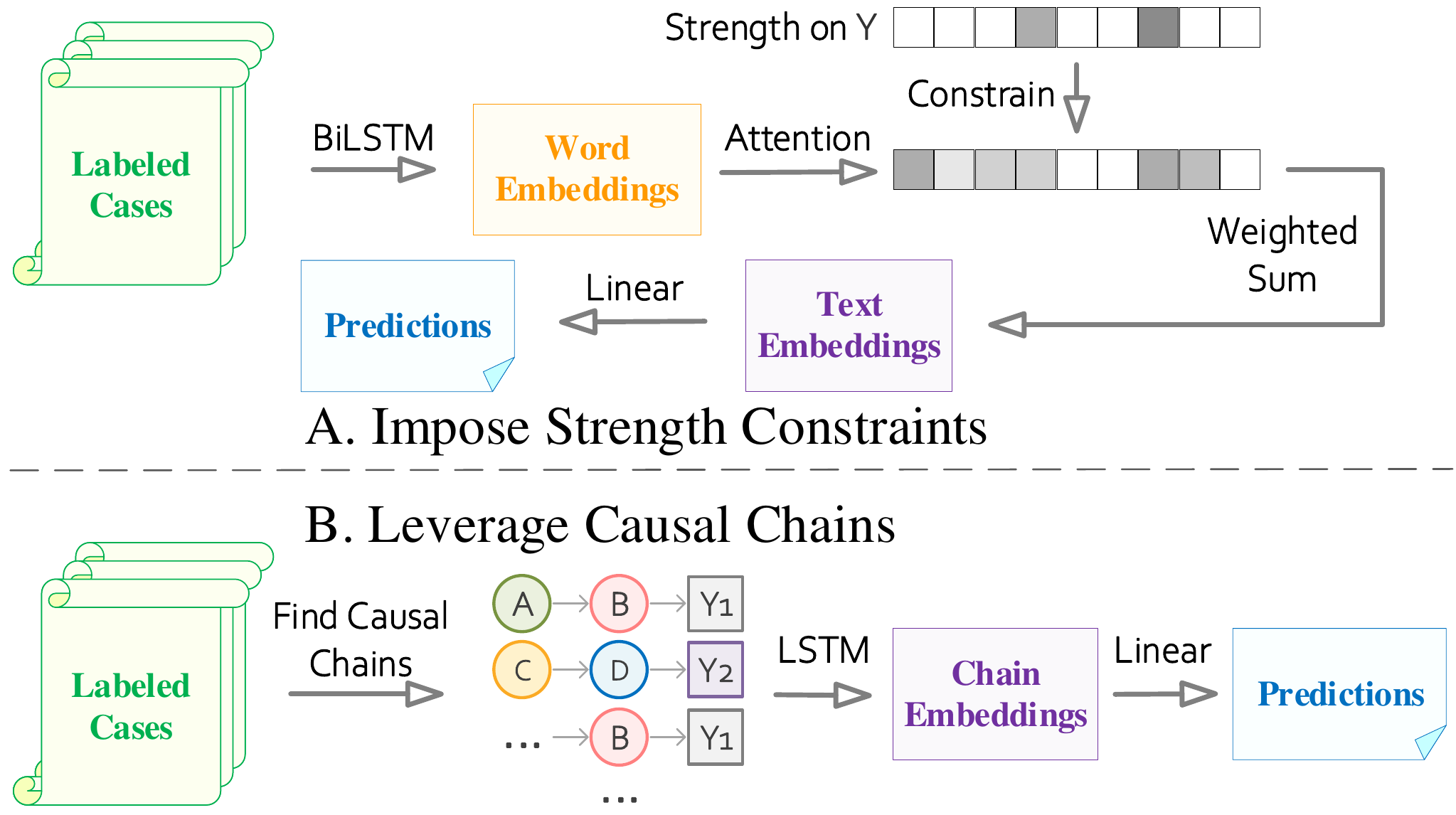}
    \caption{Two ways of integrating causal analysis and neural networks.}
    \label{fig-NN}
\end{figure}

Neural networks (NN) are considered to be good at exploring large volumes of textual data. This motivates us to integrate the causal framework with NN, to benefit each other.
Here we propose two integration methods as shown in Figure~\ref{fig-NN}.
\subsection{Imposing Strength Constraint}
\label{attconstraint}
First, we inject the estimated causal strength to constrain the attention weights of a Bi-LSTM with attention model~\cite{zhou2016attention}. A Bi-LSTM layer is first applied to the fact descriptions to obtain contextual embeddings $\mathbf{H}=\{\mathbf{h}_1,\mathbf{h}_2,\ldots,\mathbf{h}_n\}, \mathbf{h}_i \in \mathbb{R}^{b_0}$, where $b_0$ is the dimension of embeddings. Then, an attention layer assigns different weights $\{a_1,a_2,\ldots,a_n\}$ to each word, and sums the words up according to the weights to build a text embedding $\mathbf{v}$:
\begin{equation}
a_i=\frac{\mathrm{exp}(\mathbf{q}^T \cdot \mathbf{h}_i)}{\sum_{k=1}^{n} \mathrm{exp}(\mathbf{q}^T \cdot \mathbf{h}_k)}, \mathbf{v}=\sum_{i=1}^{n} a_i \times \mathbf{h}_i,
\end{equation}
where $\mathbf{q} \in \mathbb{R}^{b_0}$ is a learnable query vector. Finally, we apply two fully connected layers to the text embedding $\mathbf{v}$, and form the prediction vector $\mathbf{r}_{\textit{cons}}$.

Besides a cross-entropy loss $L_{\textit{cross}}$ on $\mathbf{r}_{\textit{cons}}$, we introduce an auxiliary loss $L_{\textit{cons}}$ to guide the attention module with the causal strength learned from \textit{GCI}. Given the golden label $c_j$, for each word $w_i$ which belongs to the factor $f$, $\tilde{\psi}_{T_f,Y_j}$ is the corresponding causal strength, and $g_i$ is the normalized strength over the whole sequence. $L_{\textit{cons}}$ is set to make the attention weights close to the normalized strength:
\begin{equation}
\begin{aligned}
& L_{\textit{cons}} = \sum_{i=1}^{n} (a_i-g_i)^2, \\ 
& L=L_{\textit{cross}}+\alpha L_{\textit{cons}}.
\end{aligned}
\end{equation}
Note that in the validation and testing stages, the inputs do not contain any strength constraint and golden charge information. Therefore, we select the epoch with the least cross-entropy loss in the validation stage to evaluate on the test set.

\subsection{Leveraging Causal Chains}
Causal chains are another type of knowledge that can be captured from causal graphs. In the legal scenario, causal chains depict the process of committing crimes. They can also be treated as the summarization of cases or behavioural patterns for each charge. 
Therefore, the second approach is to leverage the causal chains directly, as the chains may contain valuable information for judgement. For a given text, we extract factors and traverse all causal chains composed by the factors from the sampled causal graphs, $Chains=\{chain_1, chain_2, \ldots, chain_m\}$. In this task, we only consider chains ending up with treatments of charges, as they are more relevant with the judgement. An LSTM layer is applied to each chain, and all the chains are pooled to build case representation $\mathbf{c} \in \mathbb{R}^{b_0}$:
\begin{equation}
\begin{aligned}
& \mathbf{ch}_{i}=\sum_{j=1}^{l_i}(\mathrm{LSTM}(chain_{i})_j), \\
& \mathbf{c}=\mathrm{MaxPooling}(\mathrm{BIC}(G_q, \mathbf{X})\times \mathbf{ch}_{i}), \\
& \quad \quad \quad 1 \leq i \leq m,\ \ chain_i \in G_q,
\end{aligned}
\end{equation}
where $l_i$ indicates the length of $chain_i$.
The case representation $\mathbf{c}$ is then fed to two fully connected layers to make the prediction $\mathbf{r}_{\textit{chain}}$, and a cross-entropy loss is used to optimize the model.

\section{Experiments and Evaluation}
\subsection{Experimental Setup}
\textbf{Dataset.}
\begin{table}[t!]
	\centering
    \scalebox{0.95}{
	\scriptsize
	\begin{tabular}{l|c|c}
		\toprule
		\textbf{Charge Sets} & \textbf{Charges} & \textbf{\#Cases} \\
		\midrule
		\multirow{2}{*}{Personal Injury} & Intentional Injury \& Murder \&  & 6377 / 2282 / \\
		& Involuntary Manslaughter & 1989 \\
		\midrule
		\multirow{2}{*}{Violent Acquisition} & Robbery \& Seizure \&  & 5020 / 2113 / \\
		& Kidnapping & 622 \\
		\midrule
		F\&E & Fraud \& Extortion & 3536 / 2149 \\
		\midrule
		\multirow{2}{*}{E\&MPF} & Embezzlement \& & \multirow{2}{*}{2391 / 1998} \\
		& Misappropriation of Public Funds \\
		\midrule
		\multirow{2}{*}{AP\&DD} & Abuse of Power \& & \multirow{2}{*}{1950 / 1938} \\
		& Dereliction of Duty  \\
		\bottomrule
	\end{tabular}
	}
	\caption{Summary of the similar charge sets.}
	\label{table-dataset}
\end{table}
For the similar charge disambiguation task, we pick five similar charge sets from the \textit{Criminal Law of the People's Republic of China}~\citep{NPC2017Criminal}, which are hard to discriminate in practice~\citep{ouyang1999confusing}, and select the corresponding fact descriptions from the Chinese AI and Law Challenge (CAIL2018)~\citep{xiao2018cail2018}. Detailed statistics of the charge sets are given in Table~\ref{table-dataset}. 
Note we filter out the cases whose judgements include multiple charges from one charge set. The fact descriptions in our dataset are in Chinese.

\paragraph{Our Models.}
We evaluate our graph-based causal inference (\emph{GCI}) framework as described in Section~\ref{pure}, and two models integrating \emph{GCI} with NN (\emph{Bi-LSTM+Att+Cons} and \emph{CausalChain}) as described in Section~\ref{NN}.

\paragraph{Comparison Models.}
To study the effect of causal relationship captured by \emph{GCI}, we implement a variant called \emph{GCI-co}, which is built upon a correlation-based graph rather than our discovered causal graph. In detail, we compute the Pearson correlation coefficient $\phi$ for every two factors, and draw an edge if $\phi>0.5$. The direction of the edge is from the factor that appears earlier in the text more often, to the other. Then we compare \emph{GCI} and two integration methods with NN baselines, including \emph{LSTM}, \emph{Bi-LSTM} and \emph{Bi-LSTM+Att}. \emph{Bi-LSTM+Att} is a common backbone of legal judgement prediction models, while we do not add multi-task learning~\citep{luo2017learning} and expert knowledge~\citep{xu2020distinguish} for simplicity. Since the prior knowledge learned from pre-trained models may result in unfair comparison, we do not choose the models such as BERT~\citep{devlin2018bert} as baselines and backbones to eliminate the influence.
Previous works integrating text into causal inference are not able to find causal relationships inside text, so we do not take them into comparison.

We select a set of training ratios, 1\%, 5\%, 10\%, 30\%, and 50\%, to study how the performance gap changes along with different training data available. For each setting, we run the experiments on three random seeds and report the average accuracy (Acc) and macro-F1 (F1). More details about baselines, parameter selection, and training process are in Appendix~\ref{impl}.

\subsection{Main Results}
\begin{table*}[t!]
	\centering
	\scalebox{1.02}{
	\scriptsize
	\begin{tabular}{ll*{6}{|c}}
		\toprule
		\multicolumn{2}{c|}{\multirow{2}{*}{\textbf{Models}}} & \textbf{Personal} & \textbf{Violent} &  \multirow{2}{*}{\textbf{F\&E}} & \multirow{2}{*}{\textbf{E\&MPF}} & \multirow{2}{*}{\textbf{AP\&DD}} & \multirow{2}{*}{\textbf{Average}} \\
		& & \textbf{Injury} & \textbf{Acquisition} & & & \\
\midrule
\multirow{5}{*}{LSTM} & 1\% & 60.94 / 37.91 & 58.48 / 29.33 & 63.91 / 47.00 & 53.56 / 39.84 & 52.08 / 46.13 & 57.79 / 40.04 \\
& 5\% & 61.97 / 44.88 & 67.09 / 35.86 & 71.60 / 68.68 & 59.89 / 56.88 & 54.12 / 48.53 & 62.93 / 50.97 \\
& 10\% & 76.45 / 67.81 & 65.64 / 47.62 & 82.14 / 80.74 & 70.21 / 70.00 & 55.46 / 51.29 & 69.98 / 63.49 \\
& 30\% & 85.37 / 81.27 & 74.43 / 66.05 & 88.10 / 87.33 & 71.60 / 70.82 & 65.61 / 65.19 & 77.02 / 74.13 \\
& 50\% & 85.67 / 83.02 & 80.10 / 72.27 & 90.04 / 89.06 & 75.59 / 75.46 & 69.65 / 69.62 & 80.21 / 77.89 \\

\midrule
\multirow{5}{*}{Bi-LSTM} & 1\% & 62.29 / 40.81 & 53.86 / 33.25 & 62.95 / 43.27 & 54.54 / 41.91 & 48.98 / 37.84 & 56.52 / 39.42 \\
& 5\% & 74.00 / 69.52 & 65.18 / 38.99 & 60.34 / 56.96 & 61.88 / 61.63 & 51.77 / 46.23 & 62.63 / 54.66 \\
& 10\% & 76.66 / 71.86 & 67.10 / 46.07 & 85.31 / 84.37 & 60.08 / 53.34 & 60.20 / 57.95 & 69.87 / 62.72 \\
& 30\% & 85.46 / 82.53 & 75.30 / 64.12 & 87.57 / 86.58 & 70.45 / 69.64 & 65.45 / 65.12 & 76.85 / 73.60 \\
& 50\% & 87.19 / 85.01 & 78.43 / 69.94 & 90.43 / 89.83 & 76.08 / 75.78 & 71.12 / 70.50 & 80.65 / 78.21 \\

\midrule
\multirow{5}{*}{GCI} & 1\% & 69.54 / 49.77 & 57.08 / 42.55 & \textbf{82.81$^*$} / \textbf{82.56$^*$} & \textbf{74.65$^*$} / \textbf{70.22$^*$} & 62.47 / 61.72 & \textbf{69.31$^*$} / \textbf{61.36$^*$} \\
& 5\% & 81.19 / 75.58 & 69.70 / \textbf{60.39$^{\dag}$} & 88.25 / \textbf{87.24$^{\dag}$} & \textbf{83.27$^{\dag}$} / \textbf{83.06$^{\dag}$} & \textbf{78.09$^{\dag}$} / \textbf{77.95$^{\dag}$} & \textbf{80.10$^{\dag}$} / \textbf{76.84$^{\dag}$} \\
& 10\% & 80.33 / 74.50 & 74.06 / \textbf{67.31$^{\S}$} & 87.97 / 87.51 & \textbf{85.23$^{\S}$} / \textbf{84.62$^{\S}$} & \textbf{78.36$^{\S}$} / \textbf{78.31$^{\S}$} & \textbf{81.19$^{\S}$} / \textbf{78.45$^{\S}$} \\
& 30\% & 84.83 / 80.10 & 75.99 / 70.64 & 89.31 / 88.39 & \textbf{88.55$^{\ddag}$} / \textbf{88.21$^{\ddag}$} & 80.82 / \textbf{80.56$^{\ddag}$} & \textbf{83.90$^{\ddag}$} / \textbf{81.58$^{\ddag}$} \\
& 50\% & 85.72 / 81.62 & 76.31 / 71.45 & 90.41 / 89.14 & \textbf{89.01$^{\natural}$} / \textbf{88.63$^{\natural}$} & \textbf{81.01$^{\natural}$} / \textbf{80.90$^{\natural}$} & 84.49 / 82.35 \\

\midrule
\multirow{5}{*}{GCI-co} & 1\% & 67.49 / 44.43 & \textbf{63.70$^*$} / 34.64 & 75.72 / 67.60 & 69.08 / 67.20 & \textbf{64.93$^*$} / \textbf{64.41$^*$} & 68.19 / 55.66 \\
& 5\% & 76.70 / 63.94 & 67.65 / 34.35 & 86.63 / 85.81 & 82.23 / 81.86 & 73.94 / 73.77 & 77.43 / 67.95 \\
& 10\% & 68.05 / 45.37 & 69.26 / 46.39 & 85.62 / 84.41 & 81.23 / 79.64 & 74.21 / 74.05 & 75.67 / 65.97 \\
& 30\% & 77.31 / 63.45 & 70.42 / 50.94 & 81.44 / 80.54 & 85.71 / 85.20 & 74.43 / 74.28 & 77.86 / 70.88 \\
& 50\% & 79.21 / 69.37 & 70.38 / 50.78 & 79.30 / 77.58 & 84.39 / 83.72 & 74.16 / 73.99 & 77.49 / 71.09 \\

\midrule
\multirow{5}{*}{CausalChain} & 1\% & \textbf{73.20$^*$} / \textbf{60.31$^*$} & 63.60 / \textbf{44.02$^*$} & 68.01 / 52.93 & 66.97 / 56.66 & 63.13 / 62.30 & 66.98 / 55.24 \\
& 5\% & \textbf{81.99$^{\dag}$} / \textbf{76.03$^{\dag}$} & 70.57 / 59.85 & \textbf{88.64$^{\dag}$} / 87.21 & 75.13 / 74.74 & 71.75 / 70.38 & 77.62 / 73.64 \\
& 10\% & 81.21 / 74.71 & 73.50 / 66.66 & 87.59 / 86.36 & 79.75 / 79.45 & 74.43 / 74.11 & 79.30 / 76.26 \\
& 30\% & 85.61 / 81.00 & 74.93 / 67.30 & 89.10 / 88.19 & 81.63 / 81.25 & \textbf{80.90$^{\ddag}$} / 80.50 & 82.43 / 79.65 \\
& 50\% & 86.41 / 83.11 & 75.66 / 68.47 & 90.45 / 89.21 & 81.25 / 80.09 & 80.03 / 79.89 & 82.76 / 80.16 \\

\midrule
\multirow{5}{*}{Bi-LSTM+Att} & 1\% & 62.16 / 41.70 & 58.21 / 32.97 & 67.99 / 62.80 & 57.90 / 50.67 & 53.20 / 41.78 & 59.89 / 45.99 \\
& 5\% & 78.29 / 72.81 & 67.50 / 50.68 & 85.30 / 84.28 & 61.86 / 55.38 & 58.76 / 53.03 & 70.34 / 63.23 \\
& 10\% & 81.51 / 78.36 & 67.97 / 58.26 & 88.07 / 87.33 & 75.38 / 74.86 & 58.82 / 55.82 & 74.35 / 70.93 \\
& 30\% & 86.07 / 83.49 & 80.47 / 72.55 & 88.97 / 88.41 & 81.53 / 81.14 & 72.84 / 72.65 & 81.98 / 79.65 \\
& 50\% & 87.25 / 85.38 & 82.27 / 74.15 & 91.56 / 91.05 & 82.29 / 82.11 & 73.70 / 73.65 & 83.41 / 81.27 \\

\midrule
\multirow{5}{*}{\shortstack{Bi-LSTM+Att\\+Cons}} & 1\% & 70.12 / 59.46 & 54.29 / 40.34 & 78.25 / 76.80 & 61.03 / 60.62 & 53.84 / 44.93 & 63.51 / 56.43 \\
& 5\% & 79.07 / 75.89 & \textbf{73.09$^{\dag}$} / 56.84 & 86.80 / 86.35 & 66.86 / 59.89 & 72.27 / 72.18 & 75.62 / 70.23 \\
& 10\% & \textbf{83.33$^{\S}$} / \textbf{79.70$^{\S}$} & \textbf{76.26$^{\S}$} / 64.62 & \textbf{88.76$^{\S}$} / \textbf{88.02$^{\S}$} & 80.03 / 79.64 & 73.53 / 73.48 & 80.38 / 77.09 \\
& 30\% & \textbf{86.55$^{\ddag}$} / \textbf{83.85$^{\ddag}$} & \textbf{81.48$^{\ddag}$} / \textbf{73.15$^{\ddag}$} & \textbf{89.80$^{\ddag}$} / \textbf{89.35$^{\ddag}$} & 81.82 / 81.31 & 79.46 / 79.35 & 83.82 / 81.40 \\
& 50\% & \textbf{88.31$^{\natural}$} / \textbf{86.18$^{\natural}$} & \textbf{82.72$^{\natural}$} / \textbf{76.03$^{\natural}$} & \textbf{92.05$^{\natural}$} / \textbf{91.55$^{\natural}$} & 83.02 / 82.69 & 80.72 / 80.64 & \textbf{85.36$^{\natural}$} / \textbf{83.42$^{\natural}$} \\
		\bottomrule
	\end{tabular}
	}
	\caption{Performance on similar charge disambiguation. The first number is Acc and the second number is F1. Highest results are in bold, and different symbols indicate different training ratios. }
	\label{table-result}
\end{table*}

Table~\ref{table-result} reports the charge disambiguation performance of our models and comparison models.

\paragraph{Causal Graph vs. Correlation-based Graph.} 
\emph{GCI} outperforms \emph{GCI-co} by 4.5\% on average Acc, and 9.8\% on average F1, indicating the graph constructed by mining causal relations better captures the relationship between charges and factors.

\paragraph{Causal Inference vs. Neural Networks.} Comparing \emph{GCI} with NN baselines \emph{LSTM}, \emph{Bi-LSTM} and \emph{Bi-LSTM+Att}, we observe in few-shot settings (1\%, 5\%), \emph{GCI} outperforms NNs by about 10\% on average, since NNs tend to underfit in few-shot settings. 
However, with the increase of training data, the performance gap becomes narrower and consequently, NNs outperform \emph{GCI} in several cases. Compared with \emph{GCI}, NNs have the advantage of learning from large amounts of unstructured data.

\paragraph{Adding Strength Constraints.} We can see that \emph{Bi-LSTM+Att+Cons} outperforms \emph{Bi-LSTM+Att} by around 1-5\%. The performance gap is much larger in few-shot settings. This suggests that our estimated causal strength is helpful for attention-based models to capture the key information in the text. 

\paragraph{Causal Chains vs. Whole Text.} 
Both \emph{CausalChain} and \emph{LSTM} are a straightforward application of unidirectional LSTM, but over different texts, one for our extracted causal chains and the other for the whole fact description.  
 We find \emph{CausalChain} outperforms \emph{LSTM} by 8.2\% on average Acc and 11.7\% on average F1. 
The difference shows that causal chains contain condensed key information that contributes to the judgement, while the whole description may contain far more irrelevant information that may disturb the prediction. We also conduct experiments on combining causal chains and the whole plain text, but simply concatenating them does not work well since the whole text may introduce noise, and better integration methods are needed, which we leave for future work. 
\section{Analysis}
\begin{figure}[t!]
    \centering
    \includegraphics[width=0.95\linewidth]{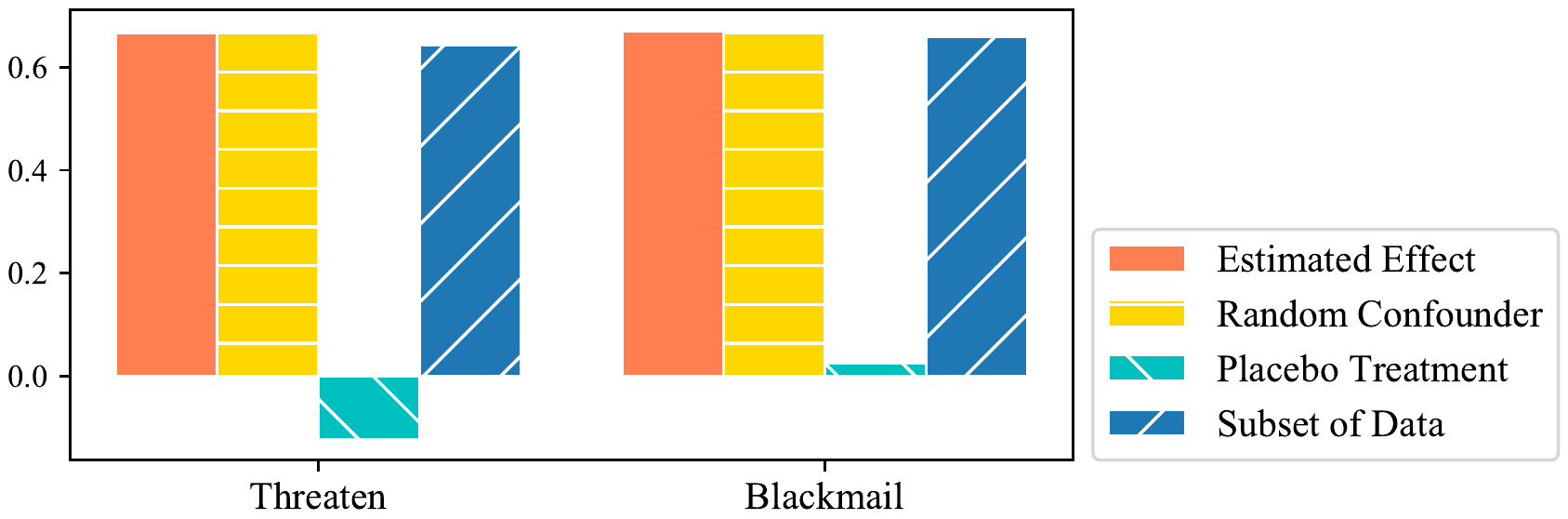}
    \caption{An example of sensitivity analysis for treatments of \texttt{extortion} in the causal graph of F\&E.}
    \label{fig-refuter}
    \vspace{-5mm}
\end{figure}

\subsection{Quality of Causal Graphs}
To analyze the robustness of the causal discovery process, we apply sensitivity analysis on the causal graphs. In detail, we make disturbance to the original causal relations, and examine the sensitivity of causal effect towards the violations.
Following~\citet{kiciman2018tutorial}, we use three refuters for examination: 1) \textit{Random Confounder}, a new confounder with random value is added to the graph, and ideally, the causal strength should remain the same as before. 2) \textit{Placebo Treatment}, the value of a treatment is replaced by a random value, so the treatment becomes a placebo, and the strength should be zero. 3) \textit{Subset of Data}, we use a subset of cases to recalculate the strength. Ideally, the strength estimation will not vary significantly. 

We take a sampled causal graph of F\&E (Fraud \& Extortion) as an example, and exhibit the refuters on the treatments of charge \texttt{extortion} in Figure~\ref{fig-refuter}. Causal strength is almost the same as before after \textit{Random Confounder} and \textit{Subset of Data} refutation; and turns to nearly zero after \textit{Placebo Treatment}. The results show that our graph construction method is robust against disturbance.

\subsection{Causal Chains in Graph}
\label{sec-chain}
\begin{figure}
    \centering
    \includegraphics[width=0.9\linewidth]{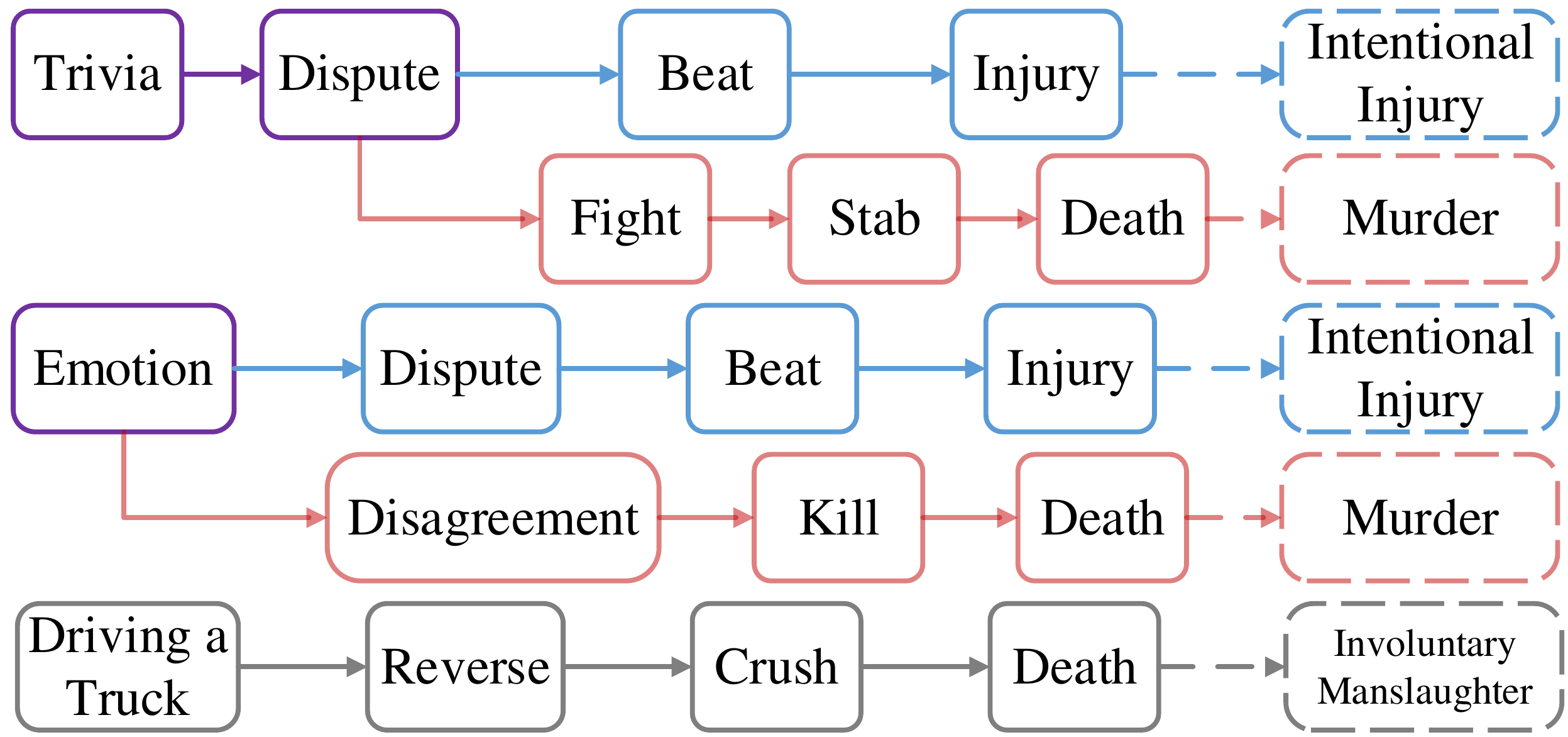}
    \caption{Causal chains in Personal Injury's graph.}
    \label{fig-chains}
\end{figure}
Causal chains manifest how effective \emph{GCI} is in terms of inducing common patterns of suspect's behaviours. It is helpful for people to better understand the core part of legal cases. Here we select the causal graph of Personal Injury (Intentional Injury \& Murder \& Involuntary Manslaughter) and showcase several causal chains underlying the text. As shown in Figure~\ref{fig-chains}, the chains depict common patterns of these charges, from the initial causes, to the final criminal behaviour. More examples are provided in Appendix~\ref{showchain}.

Now the question is how the graph structures help to discriminate similar charges. Here we analyze the nuance between the causal chains of two similar charges E\&MPF (Embezzlement \& Misappropriation of Public Funds).
For both charges, the cases often first give the background that someone held a certain position of authority, thus had power. Then both kinds of cases describe that the person illegally obtained a large amount of money by utilizing his power. While the cases in E\&MPF share very similar context, there exists nuance between them: embezzlement emphasizes that someone privately and illegally possesses the bulk of money, while misappropriation of public funds emphasizes that someone would temporally use the money for a certain purpose. By observing the causal chains of two charges, \emph{GCI} could capture the slight difference well: for embezzlement, the causal chains tend to be \textit{work / take charge of $\rightarrow$ take advantage of (position/power)}; for misappropriation of public funds, the causal chains tend to be \textit{take charge of $\rightarrow$ take advantage of (position/power) $\rightarrow$ misappropriate $\rightarrow$ profit}. The difference between the former and latter chains is whether the person had subsequent behaviour (e.g., using the money for purposes like making profits). We could observe that the difference in causal chains accords with the definitions of the two charges. 

\subsection{Effect of Integrating Causal Strength with Attention}
\label{sec-humanevaluation}
Following~\citet{lei2017interpretable}, we conduct human evaluation on words accorded with high attention weights and compare the evaluation results of standard (\emph{Bi-LSTM+Att}) and constraint-based attention models (\emph{Bi-LSTM+Att+Cons}).

\begin{table}[t!]
	\centering
	\scriptsize
    \begin{tabular}{l*{2}{|c}}
		\toprule
		\textbf{Charge Sets} & \emph{Bi-LSTM+Att} & \emph{Bi-LSTM+Att+Cons} \\
		\midrule
		Personal Injury & 3.03 & \textbf{3.17} \\
		Violent Acquisition & 3.18 & \textbf{3.74} \\
		F\&E & 3.34 & \textbf{3.65} \\
		E\&MPF & 3.13 & \textbf{3.27} \\
		AP\&DD & 3.08 & \textbf{3.13} \\
		\bottomrule
	\end{tabular}
	\caption{Results of human evaluation. Better results are in bold.}
	\label{table-human}
\end{table}

For each set of charges, we train both models with 10\% data, and randomly select 30 cases that both models predict correctly. For the total 150 cases, we showcase content, charge names, attention weights above 0.05, and corresponding words. Each participant is asked to score from 1 to 5 for the extent of how beneficial the extracted keywords are to disambiguation. A higher score means that the attention weights indeed capture the keywords. Each case is assigned to at least four participants. Results are shown in Table~\ref{table-human}. We observe that the constraint-based model is better at explanation than normal attention-based models on all five charge groups. Take Violent Acquisition as an example. Although the cases are predicted correctly by \emph{Bi-LSTM+Att}, the model tends to attend to words \textit{bag}, \textit{RMB} and \textit{value}, which frequently occur but cannot be treated as clues for judgement. Instead, \emph{Bi-LSTM+Att+Cons} values factors such as \textit{grab}, \textit{rob} and \textit{hold}, which are more helpful for judgement.


\section{Related Works}
\label{relatedworks}
\paragraph{Causal Inference with Text.} Recently, a few works try
to take text into account when performing causal inference. \citet{landeiro2016robust} introduce causal inference to text classification, and manage to remove bias from certain out-of-text confounders. \citet{wood2018challenges} use text as a supplement of missing data and measurement error for the causal graphs constructed by structured data. \citet{egami2018make} focus on mapping text to a low-dimensional representation of the treatment or outcome. \citet{veitch2019using} and \citet{yao2019estimation} treat text as confounder and covariate, which help to make causal estimation more accurate. These works all build causal graphs manually, and regard text as a whole to be one of the factors. 
In contrast, we set our sights on text containing rich causal information in itself.
\citet{paul2017feature} looks into text by computing propensity score for each word, but only focuses on causal relationship between words and sentiment. We instead take a causal graph perspective, discover and utilize causal relationship inside text to perform reasoning.

\paragraph{Neural Networks for Causal Discovery.}
Recently, researchers attempt to apply neural networks to causal discovery~\cite{ponti-korhonen-2017-event,alvarez-melis-jaakkola-2017-causal,ning-etal-2018-joint,gao-etal-2019-modeling,Weber2020CausalIO}. However, \citet{alvarez-melis-jaakkola-2017-causal} model causal relationship by correlation, which may introduce bias into causal inference; \citet{ning-etal-2018-joint} and \citet{gao-etal-2019-modeling} merely focus on capturing causality by explicit textual features or supervision from labeled causal pairs. 
There are also a line of works focusing on how to use neural networks to summarize confounders and estimate treatment effects \cite{DBLP:conf/nips/LouizosSMSZW17,DBLP:conf/nips/Yao0LHGZ18,DBLP:journals/corr/abs-1808-07804}, which are parts of the whole causal inference process. 
\citet{Weber2020CausalIO} show how to formalize causal relationships in script learning, but it is limited to pairwise learning of events and cannot be generalized to sequential and compositional events. 

\paragraph{Legal Judgement Prediction.}
Previous works in legal text analysis focus on the task of legal judgement prediction (LJP). \citet{luo2017learning} and \citet{zhong2018legal} exploit neural networks to solve LJP tasks. \citet{zhongiteratively} provide interpretable judgements by iteratively questioning and answering. Another line pays attention to confusing charges: \citet{hu2018few} manually design discriminative attributes, and \citet{xu2020distinguish} use attention mechanisms to highlight differences between similar charges. Using knowledge derived from causal graphs, \textit{GCI} exhibits a different and interpretable discrimination process.
\section{Discussion}

Although \emph{GCI} is effective on its own and when working with powerful neural network models, 
there is still room for further improvement.
\paragraph{More Precise Causal Inference Models.} The causal estimation results of \emph{GCI} is based on the constructed causal graphs in former stages, and the automated construction process may bring imprecise factors and even omissions. As clustering algorithms try to summarize the general characteristics of text, descriptions with subtle differences may be clustered into one factor, but the differences matter in legal judgement. For example, in Personal Injury’s graph, different ways of killing are summarized as the factor \textit{kill}, therefore lose valuable information. Specifically, beaten to death might occur in cases of involuntary manslaughter, while shooting cases are more likely to be associated with murder.
Also, factors with low frequency may be omitted in clustering, but are actually useful for discrimination.
Overall, under the circumstance without much expert effort, it is worthwhile to explore how to construct a more reliable causal graph. 



\paragraph{Deep understanding on legal documents.} Although \emph{GCI} to some extent tackles the challenges of incapability of causal inference methods on unstructured text, it may make mistakes when facing complex fact descriptions. \textit{Negation semantics} is a typical example. It is occasional to see negation word usage in fact descriptions, which usually indicates that someone did \textit{not} have a certain behaviour. However, \emph{GCI} has not considered this aspect, and may be awry in cases containing negation usage. Besides, \textit{pronoun resolution} is also an important aspect that may confuse models. For example, certain behaviour is done by the victim and the subject of the behaviour is a pronoun. If the model was unaware of the subject of the behaviour, it would be counted as criminal's behaviour and introduces noise to later inference stage. Moreover, \textit{intent} could be a decisive factor when discriminating charges, for example, murder and involuntary manslaughter. But it may not be mentioned explicitly in the fact descriptions. It should be better to recover the complete course of fact and recognize the implicit intents between the lines with deep understanding of the context and relevant common sense knowledge. 


\section{Conclusions}
We propose \textit{GCI}, a graph-based causal inference framework to discover causal information in text, and design approaches to integrate causal models and neural networks. 
In the similar charge disambiguation task, we show our approaches capture important evidence for judgement and nuance among charges. Further analysis demonstrates the quality of causal graphs, value of causal chains, and interpretability of computed causal strength. 

\section{Ethical Considerations}
\label{ethics}
\subsection{Intended Use}
We aim to facilitate legal service with our proposed \textit{GCI}, providing valuable evidence instead of directly making judgements. We hope that legal AI models could assist legal workers in underdeveloped areas, helping them to explore key points in cases, discriminate from similar crimes, and make better decisions. By treating like cases alike in different regions of the country, influences of judges' intellectual flaws and arbitrariness are weakened, and rights of people, both the defendants and the victims, are protected.
\paragraph{Failure Mode.} 
The model may give wrong evidence in some cases, but this will not cause significantly bad impact. The process of GCI is transparent. Extracted factors, the causal relations between them, and causal chains that lead to the final decision are all shown to users. By checking these ``rationales'' of model reasoning, users can clearly find where goes wrong, and not adopt the outputs, or intervene and correct the model.
\paragraph{Misuse Potential.}
We emphasize that such a model cannot be used individually, as the trial process is seriously conducted and regulated by the judicial system. In the actual judicial process, the prosecutors, judges, and lawyers are under strict supervision. We do not think there is a possibility for them to misuse computer models.
\subsection{Bias Analysis}
Criminal behaviour is very unbalanced in gender. Take the three charges in our Personal Injury charge set as an example. \citet{lin2020bluebook} counted gender ratio in criminal cases from 2013 to 2017 in China, and the ratios of male defendant are 94.70\% (Intentional Injury), 87.72\% (Murder), and 93.97\% (Involuntary Manslaughter). The disparity in defendants of male and female leads to the small proportion of female cases in training corpus. Therefore, female cases may be inadequately trained. If this results in more incorrect predictions for female cases, women's rights are violated. 

Following~\citet{dixon2018measuring} and~\citet{park2018reducing}, we use False Positive Equality Difference (FPED) and False Negative Equality Difference (FNED) to examine the performance difference in two genders. They are defined as:
\begin{equation}
    \begin{aligned}
    & \textrm{FPED}=\sum_{t \in T}|\textrm{FPR}-\textrm{FPR}_t|, \\
    & \textrm{FNED}=\sum_{t \in T}|\textrm{FNR}-\textrm{FNR}_t|,
    \end{aligned}
\end{equation}
where FPR is false positive rate of classification, FNR is false negative rate, and $T=\{male, female\}$. The two metrics quantify the extent of variation between the performances of two genders. 
\begin{table}[t!]
	\centering
	\footnotesize
    \begin{tabular}{l*{2}{|c}}
		\toprule
		\textbf{Metrics} & \emph{Bi-LSTM+Att} & \emph{Bi-LSTM+Att+Cons} \\
		\midrule
		FPED & 0.048 & \textbf{0.032} \\
		FNED & 0.065 & \textbf{0.049} \\
		\bottomrule
	\end{tabular}
	\caption{Results of equality difference. Better results are in bold.}
	\label{table-ed}
\end{table}

Applying them to \textit{Bi-LSTM+Att} and \textit{Bi-LSTM+Att+Cons} models in Personal Injury charge set, the results are shown in Table~\ref{table-ed}. The model with causal constraints achieves smaller variance measured by both metrics, which reduce between $1/4$ to $1/3$ of the unfairness in performance of \textit{Bi-LSTM+Att}. This shows the superiority of our model with causal knowledge. 
Compared with normal neural networks, our constraint-based model utilizes causal relations, which are more stable to the number of occurrences.

Though adding causal knowledge narrows the equality difference, it still exists in \textit{Bi-LSTM+Att+Cons} (the metrics are greater than zero). Other types of bias may also exist in our model, given that the training corpora contain decisions of humans and systemic bias of humans may be preserved. Further debiasing method is needed if the model is put into real use.

In general, we believe that adding causal knowledge to decision making will help debiasing, and the transparent exhibition of causal graphs and chains will enable people to find biases in time and correct them.

\section*{Acknowledgments}
This work is supported in part by National Hi-Tech R\&D Program of China (2018YFC0831900). We would like to thank the anonymous reviewers for the helpful discussions and suggestions. Also, we would thank Yuxuan Lai, Liunian Harold Li, Jieyu Zhao, Chen Wu and Xingchen Lan for advice about experiments and writing. For any correspondence, please contact Yansong Feng.

\bibliography{naacl2021,anthology,custom}
\bibliographystyle{acl_natbib}

\clearpage
\appendix
\section{GFCI Algorithm}
\label{algo}
GFCI~\citep{ogarrio2016hybrid} is a combination of a constraint-based causal discovery algorithm FCI~\citep{spirtes2013causal} and a score-based algorithm FGES~\citep{ramsey2016million}. Based on the skeleton of FCI, it uses initialization from FGES to improve accuracy and efficiency.

FCI takes sample data and optional background knowledge as input, and guarantees to represent the Markov equivalence class of the true causal DAG. It has two phases: adjacency and orientation. It does not rely on no latent confounder assumption, but it performs relatively poorly, especially on real data.

On the other hand, FGES greedily searches over potential DAGs, and outputs the highest scoring graph it finds. It is fast and accurate when its assumptions are satisfied, but it relies on the condition that there are no latent confounders.

GFCI takes the output of FGES as an initialized graph, and the graph is further augmented by FCI's adjacency phase, in which some adjacencies are removed by conditional independence tests. A similar process happens in the orientation phase. Orientations of FGES are provided as initialization, and further orientations from FCI are applied. 
Proof in~\citet{ogarrio2016hybrid} guarantees the GFCI algorithm outputs a PAG that represents the true causal DAG. 

\begin{figure}[t!]
    \centering
    \includegraphics[width=\linewidth]{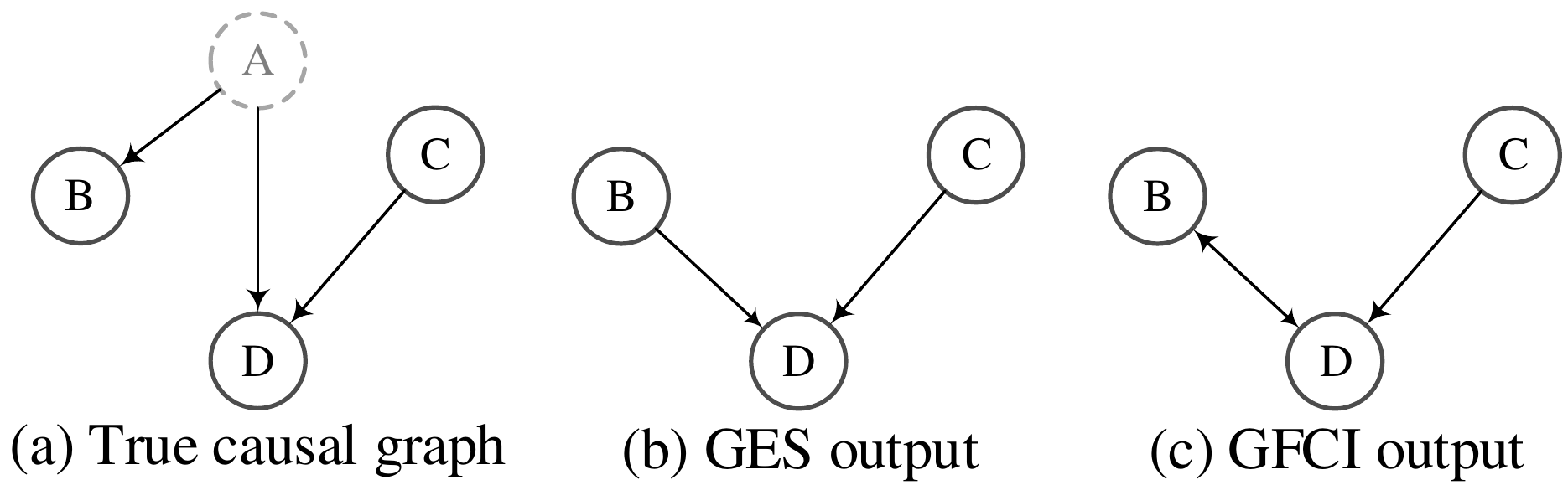}
    \caption{Comparison between the output of GES and GFCI. $A$ is a latent variable, which is also an unobserved confounder of $B$ and $D$. This implies that in fact there is no causal relationship between $B$ and $D$ but without the consideration of unobserved confounders, GES tends to be awry in this situation.}
    \label{fig-pag}
\end{figure}

Figure~\ref{fig-pag} shows the advantage of GFCI. When factor $A$ is unobserved, score-based algorithms like GES~\citep{chickering2002optimal} will wrongly discover edge $B \rightarrow D$, while GFCI correctly recognizes the relation between factors $B$ and $D$: there is an unobserved confounder of $B$ and $D$.

\section{Implementation Details}
\label{impl}
When the training set ratio is 1\%, we select $p=15$ keywords for each charge, and cluster the keywords of both charges into $q=20$ factors; for the training set ratio 5\% and 10\%, $p=25$ and $q=30$; for the training set ratio 30\% and 50\%, $p=40$ and $q=60$. We use K-means~\citep{macqueen1967some} for clustering.
When learning causal relationship, besides cases in the training set, we also use some unlabeled cases of other charges to improve the interpretability of the causal graph. Note that removing these cases will not hurt the performance. We sample $Q=5$ causal graphs for each PAG.

For the neural network based models (both the baselines and our proposed models), 10\% of the training set is used as the validation set. The models are trained for 30 epochs, and the early stopping mechanism takes effect when the validation loss does not drop for more than 1000 batches. We use Adam optimizer with learning rate 0.001, and the dropout rate is 0.5. 
We use Tencent AILab Chinese Word Embedding~\citep{song2018directional} as word embeddings, and the dimension of each word is 200. The hidden sizes are $b_0=128$ and $b_1=64$. The batchsize is selected from $\{4,8,16,32,64,128\}$ considering the dataset size and training set ratio. For \emph{Bi-LSTM+Att+Cons}, $\alpha$ is manually tuned within $\{0.1, 0.25, 0.5, 1\}$, selecting the one with best validation F1.
All the hyperparameters are empirically selected and kept the same for different models in the same dataset and the same training set ratio.

To reduce the impact of data imbalance between charges, we apply data augmentation to smaller charges whose cases are 3 times fewer than the biggest charge in that set, regardless of the training ratio. We did not keep them 1:1, in order to reflect their distributions in the real world.

The models are trained on Intel Xeon CPU. It takes about one hour to construct the causal graphs, and a few minutes to perform inference.

\section{Example of Generated PAG}
\begin{figure}
    \centering
    \includegraphics[width=\linewidth]{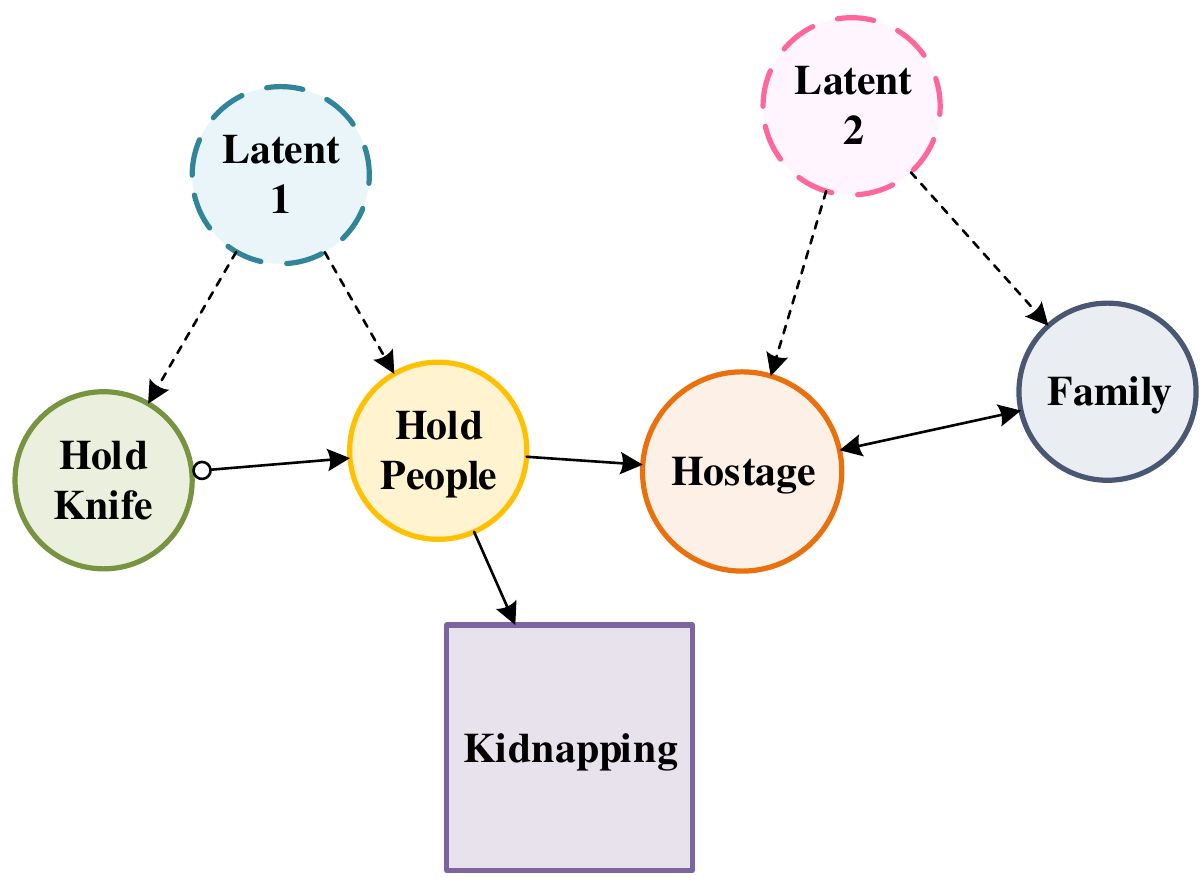}
    \caption{An example of part of the generated PAG of charge \texttt{kidnapping} in charge set Violent Acquisition. \textit{Latent1} and \textit{Latent2} are unobserved variables that do not exist in the generated PAG. Detailedly, \textit{Latent1} may denote the suspect's intent to commit a crime; and \textit{Latent2} may denote the suspect's intent to obtain properties from the hostage.}
    \label{fig-pagexample}
\end{figure}

\begin{table*}
	\centering
	\small
	\begin{tabular}{l|l}
		\toprule
		\textbf{Charge Sets} & \textbf{Chains} \\
		\midrule
		\multirow{3}{*}{Personal Injury} & trivia $\rightarrow$ dispute $\rightarrow$ beat $\rightarrow$ injury $\rightarrow$ \texttt{intentional injury} \\
		& emotion $\rightarrow$ disagreement $\rightarrow$ kill $\rightarrow$ death $\rightarrow$ \texttt{murder} \\
		& driving a truck $\rightarrow$ reverse $\rightarrow$ crush $\rightarrow$ death $\rightarrow$ \texttt{involuntary manslaughter} \\
		\midrule
		\multirow{3}{*}{Violent Acquisition} & hold a knife $\rightarrow$ violence $\rightarrow$ cash $\rightarrow$ \texttt{robbery} \\
		& necklace $\rightarrow$ pull $\rightarrow$ \texttt{seizure} \\
		& hostage $\rightarrow$ threaten $\rightarrow$ ask for $\rightarrow$ ransom $\rightarrow$ \texttt{kidnapping} \\
		\midrule
		\multirow{3}{*}{F\&E} & approach $\rightarrow$ trust $\rightarrow$ lie $\rightarrow$ \texttt{fraud} \\
		& forge $\rightarrow$ defraud $\rightarrow$ \texttt{fraud} \\
		& relationship $\rightarrow$ nude photos $\rightarrow$ threaten $\rightarrow$ \texttt{extortion} \\
		\midrule
		\multirow{4}{*}{E\&MPF} & work $\rightarrow$ take advantage of (position/power) $\rightarrow$ \texttt{embezzlement} \\
		& falsely report $\rightarrow$ arbitrage $\rightarrow$ \texttt{embezzlement}\\
		& take charge of $\rightarrow$ take advantage of (position/power)$\rightarrow$ misappropriate $\rightarrow$ profit\\
		& $\rightarrow$ \texttt{misappropriation of public funds} \\
		\midrule
		\multirow{3}{*}{AP\&DD} & (use) position $\rightarrow$ (provide) convenience $\rightarrow$ \texttt{abuse of power} \\
		& complete understand $\rightarrow$ violate $\rightarrow$ \texttt{abuse of power} \\
		& ignore $\rightarrow$ result in (bad things) $\rightarrow$ \texttt{dereliction of duty} \\
		\bottomrule
	\end{tabular}
	\caption{Exhibition of causal chains for each charge set.}
	\label{table-chains}
\end{table*}
\label{appendix-exp}
In this section, we further show part of a Partial Ancestor Graph (PAG) generated by our \emph{GCI} framework. The example is shown in Figure~\ref{fig-pagexample}. 

For the edge between \emph{Hold People} and \emph{Hostage}, the arrow is $\rightarrow$, indicating there is a causal relationship between the factors \emph{Hold People} and \emph{Hostage} (the person who was held might become a hostage). 

For the edge between \emph{Hold Knife} and \emph{Hold People}, the arrow is $\circ\!\!\rightarrow$, which means there \emph{might} be an unobserved confounder which causes the two factors, and there \emph{might} exist causal relationship between the two factors. In the first possible situation, the confounder could be \emph{Latent1}, the intent to commit a crime. A person who intends to commit a crime may hold a knife and hold people. And in the second possible situation, the person who holds a knife is likely to hold people. Therefore, there is uncertainty in predicting the causal relationship.

For the edge between \emph{Hostage} and \emph{Family}, the arrow is $\leftarrow\!\!\rightarrow$, which means there \emph{is} an unobserved confounder which causes the two factors, and there \emph{does not} exist causal relationship between them. For example, the confounder could be \emph{Latent2}, the intent to obtain properties from the hostage. The intent could not only make the suspect grab the hostage's money, but also contact families to ask for ransom. 

\section{Exhibition of More Causal Chains}
\label{showchain}

Table~\ref{table-chains} showcases three chains for each charge set. They depict common patterns of behaviours of defendants that are charged with the crime at the tail of the chains. Note that each chain merely describes one possibility and one aspect of the criminal behaviour, and the chain itself may not be sufficient to initiate such a lawsuit. For example, the chain \textit{complete understand $\rightarrow$ violate $\rightarrow$} \texttt{abuse of power} of charge set AP\&DD shows that the litigant deliberately violated the rules, 
but he/she will be charged with abuse of power only when his/her action results in major loss of public property.

\end{document}